\documentclass[10pt, a4paper]{article}
\usepackage{lrec2022} 
\usepackage{multibib}
\newcites{languageresource}{Language Resources}
\usepackage{graphicx}
\usepackage{tabularx}
\usepackage{soul}

\usepackage{enumitem}
\usepackage{amssymb}
\usepackage{comment}
\usepackage{todonotes}

\usepackage{titlesec}
\titleformat{\section}{\normalfont\large\bfseries\center}{\thesection.}{1em}{}
\titleformat{\subsection}{\normalfont\SmallTitleFont\bfseries\raggedright}{\thesubsection.}{1em}{}
\titleformat{\subsubsection}{\normalfont\normalsize\bfseries\raggedright}{\thesubsubsection.}{1em}{}
\renewcommand\thesection{\arabic{section}}
\renewcommand\thesubsection{\thesection.\arabic{subsection}}
\renewcommand\thesubsubsection{\thesubsection.\arabic{subsubsection}}

\usepackage{epstopdf}
\usepackage[utf8]{inputenc}

\usepackage{hyperref}
\usepackage{xstring}

\usepackage{color}

\title{Estimating Confidence of Predictions of Individual Classifiers and Their Ensembles for the Genre Classification Task}

\name{Mikhail Lepekhin, Serge Sharoff} 

\address{MIPT, University of Leeds \\
         Moscow, UK \\
         lepehin.mn@phystech.edu, S.Sharoff@leeds.ac.uk}

\abstract{Genre identification is a subclass of non-topical text classification. The main difference between this task and topical classification is that genres, unlike topics, usually do not correspond to simple keywords, and thus they need to be defined in terms of their functions in communication. Neural models based on pre-trained transformers, such as BERT or XLM-RoBERTa, demonstrate SOTA results in many NLP tasks, including non-topical classification. 
However, in many cases, their downstream application to very large corpora, such as those extracted from social media, can lead to unreliable results because of dataset shifts, when some raw texts do not match the profile of the training set. To mitigate this problem, we experiment with individual models as well as with their ensembles. To evaluate the robustness of all models we use a prediction confidence metric, which estimates the reliability of a prediction in the absence of a gold standard label. We can evaluate robustness via the confidence gap between the correctly classified texts and the misclassified ones on a labeled test corpus, higher gaps make it easier to improve our confidence that our classifier made the right decision. Our results show that for all of the classifiers tested in this study, there is a confidence gap, but for the ensembles, the gap is bigger, meaning that ensembles are more robust than their individual models.
\\ \newline  }

\begin{document}

\maketitleabstract

\section{Introduction}
Non-topical text classification includes a wide range of tasks aimed at predicting a text property that is not connected directly to a text topic. For example, predicting a text style, politeness, difficulty level, the age or the first language of its author, etc. Automatic genre identification \cite{santini10genreintro} is one of the standard problems of non-topical text classification. It is applied in many areas such as information retrieval, language teaching or linguistic research.

 Compared to topical text classification, genre classification has additional difficulties. First of all, the concept of genre is more complex than that of a topic. Every topic has a set of keywords and, therefore, one can define whether a text belongs to a topic or not based on the occurrence of the keywords in the text. Genre is a functional dimension of texts and in most cases, it cannot be defined just by a set of its keywords. Besides, genre typologies are often very large, while the training datasets offer few examples per more specific genres, which makes the genre harder to be classified by neural models.

Text genre identification has a long history of research. \cite{sharoff10lrec} is one of the first works on text genre classification containing a comparison of various datasets, models, and data features. It shows how traditional ML models such as SVM with various hyper-parameters and features performs for the genre classification of texts. Since then, traditional ML models have been superseded by neural models. \cite{bert} introduced BERT -- (Bidirectional Encoder Representations from Transformers), an efficient language representation model based on the Transformer architecture \cite{transformer}. It achieves state-of-the-art results for various NLP tasks, including text classification. XLM-RoBERTa \cite{xlm_roberta} is an improved variant of BERT. It has a similar architecture but uses a bigger and more genre-diverse corpus based on Common Crawl (instead of Wikipedia for the multilingual BERT). Therefore, we choose XLM-RoBERTa as the classifier for the experiments in our research.

One of the most significant problems in genre classification concerns topical shifts, according to \cite{petrenz10}. If a topic is more frequent in the training corpus for a given specific genre, then a classifier tends to predict the genre by the keywords of the topic, for example, \textit{hurricane} can be associated with the genre of FAQs \cite{sharoff10lrec}. This causes numerous unreasonable mistakes in genre classification. Conversely, a topical shift within the same genre can effect the predictions. For this reason, \newcite{petrenz10} check the accuracy of the genre classifiers via testing on the datasets from different domains, and so do we in our work. However, we introduce neural methods for confidence estimation of the genre classifiers.

Ensembles has not been experimented with in the Genre identification domain. Our paper is the first one that contains such sort of research.

We take the baseline classifiers from our work \cite{genre_bertattack}. Then the following steps are conducted in this paper to investigate the properties of ensembles for the problem of genre classification and to get a genre classifier with a higher accuracy score:

\begin{enumerate}
    \item finding the optimal hyperparameters for training the genre classifiers (\autoref{subsecOptimalEpochs}),
    \item analysing the distribution of genres in Social Media corpora (\autoref{subsecBigDatasets}),
    \item measuring of the confidence of prediction for each classifier on each pair (train dataset, test dataset) in \autoref{secConfidence}.
\end{enumerate}

The tools to replicate the experiment are available.\footnote{\url{https://github.com/MikeLepekhin/GenreClassifierEnsembles}}

\section{Confidence of prediction}
\label{secConfidenceSetup}

To estimate robustness of our predictions at the inference stage, we use a \textit{confidence} metric, which equals $1 - uncertainty$, as introduced in \cite{dropout_strikes_back}.   
The key idea is that a classifier can only be confident on a text example if it predicts the correct labels for texts with \textit{similar} embeddings. If the genre predicted by a classifier is not the same for many texts in the neighbourhood of an original text, we cannot consider the classifier confident. To simulate variation in embeddings, we apply dropout with probability 0.1 to all of the model layers, including the embedding layer and the final dense classification layer \cite{sun19}. A softmax classifier returns a probability distribution of the possible text labels. Normally, the label with the maximal probability is considered the answer of a classifier. Application of dropout to the classifier disturbs the probability distribution. We perform dropout $n$ times, and thereby we generate the corresponding probability distributions $p_1, ..., p_n$. Then we pool them into single distribution $\widehat{p}$. The maximum value of probability in $\widehat{p}$ is called \textit{the confidence of prediction}. The intuition of this metric is that if a classifier is confident in the predicted text label, it is unlikely to distribute the likelihood on other class labels significantly often. In our study, we use $n=10$ - the same as used by the authors of \cite{dropout_strikes_back}, since this provides a good balance between assessing the confidence value and the speed of computation. 

While the method is applicable to unlabeled datasets, it is beneficial to estimate the reliability of how it deals with the predicted labels. Therefore, we apply this procedure to the test dataset to compare the difference in the confidence of correct and incorrect predictions made by different classifiers. We denote this difference as \textit{confidence delta} and in our study we test the hypothesis that the value of delta helps to understand whether the text is classified correctly or not. For a given classifier model, the higher the delta, the more reliable threshold we can choose to cut off misclassified texts. It means that even in the absence of a gold-standard label, we can use the prediction confidence to say whether a classifier is likely to predict the text genre correctly or not.

\section{Training data}

\begin{table*}[!t]
\centering
\begin{tabular}{llp{0.3\linewidth}rrrr}
Genre short & Genre label & Prototypes & \multicolumn{2}{c}{LJ} & \multicolumn{2}{c}{FTD} \\
            &   &  & train &  test  & train & test \\
\hline
A1 & Argument 	& Argumentative blogs or opinion pieces 	& 1858 	& 599 & 207 & 77 \\
A4 & Fiction 	& Novels, myths, songs, film plots 	& 698	& 232 & 62 & 23 \\
A7 & Instruction 	& Tutorials, FAQs, manuals 	& 1617	& 478 & 59 & 17 \\
A8 & News    	& Reporting newswires 	& 2255 	& 787 & 379 & 103 \\
A9 & Legal   	& Laws, contracts, terms\&conditions 	& 17	& 12 & 69 & 13 \\
A11 & Personal 	& Diary entries, travel blogs 	& 2291 	& 709 & 126 & 49 \\
A12 & Promotion 	& Adverts, promotional postings 	& 195  	& 61 & 222 & 85 \\
A14 & Academic 	& Academic research papers 	& 34 	& 10 & 144 & 49 \\
A16 & Information 	& Encyclopedic articles, definitions, specifications 	& 695 & 221 & 72 & 33 \\
A17 & Review  	& Reviews of products or experiences 	& 681	& 219 & 107 & 34 \\
\hline
  & \textbf{Total} & & 10341 & 3288 & 1447 & 483 \\
\end{tabular}
\caption{Training and testing corpora \label{tabTraining}}
\end{table*}

In our experiments, we use two Russian datasets annotated with the same set of genre labels -- FTD genre dataset \cite{sharoff18genres} and texts from LiveJournal which are labeled manually for this paper. 
The first one has nearly 2000 labeled texts. The LiveJournal corpus is much bigger, it consists of more than 10500 texts see \autoref{tabTraining}.

LiveJournal is a Russian blogging platform. It consists mostly of posts and comments, sharing news, describing personal experiences or opinions. FTD is a collection of documents from multiple sources, such as Wikipedia, Reddit, and online newspapers. These two corpora are quite different in terms of genre distribution. FTD is more or less balanced when the LJ corpus is not. The legal and academic texts are especially underrepresented in LJ. It means that we have to treat the results of the classification of the Legal texts in the LJ corpus as unreliable. The largest category in both LJ and FTD concern news reporting. We collect the text from LiveJournal by random sampling of the authors and their posts. Each text in the LiveJournal dataset is labeled by two assessors. Those texts for which the labels did not coincide were additionally discussed.

Our intention is to test our genre classifiers on a domain that is different from that of the training data similarly to \cite{petrenz10}. We conduct experiments, training models on LiveJournal and FTD separately, to show how the selection of the training corpus affects metrics on the testing data, as well as on their concatenation. We split the corpora in the same ratio. For each corpus, 75\% is used for training, 25\% - for testing.

\begin{figure*}[!t]
\centering
  \includegraphics[width=.9\linewidth]{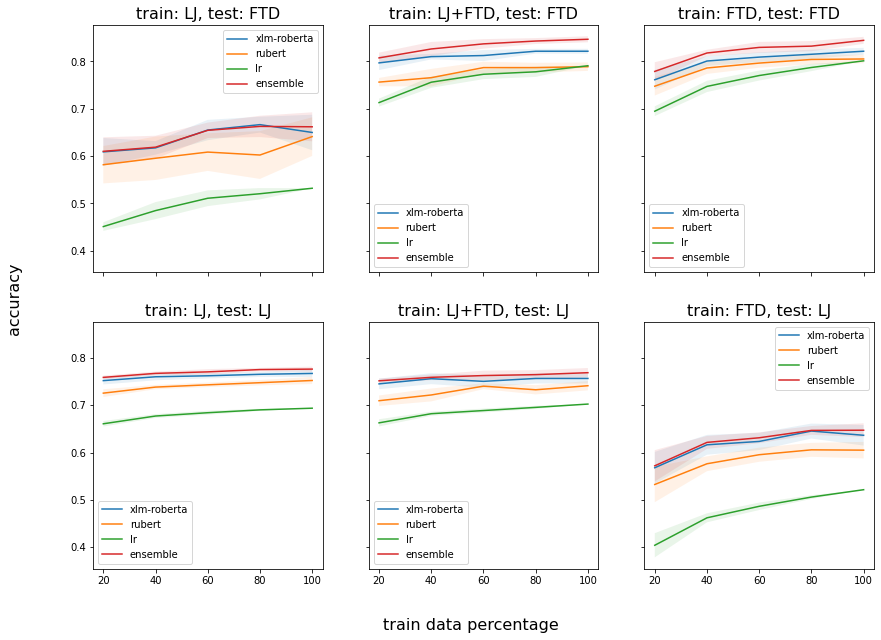}
  \caption{Dependence of the model accuracy on the train data size}
  \label{dependenceOnTrainSize}
\end{figure*}

\section{Baseline Classifiers}
\label{secBase}
\subsection{Train XLM-RoBERTa, RuBERT and Logistic Regression}

We fine-tune base RuBERT pretrained by DeepPavlov \cite{kuratov19}, base XLM-RoBERTa \cite{xlm_roberta} and a Logistic Regression classifier with character- and word-based features. In some experiments, we also trained SVM and Random Forest classifiers on the same features to compare their performance with Logistic Regression.

We use XLM-RoBERTa because it achieves the best accuracy on the Russian section of the XNLI corpus. We also used the RuBERT model because it attained the highest accuracy on the RuSentiment classification dataset among all the monolingual models for the Russian language. In each experiment, we use the Adam optimiser and $learning\_rate = 5 \cdot 10^{-5}$ as advised in \cite{sun19}. But our computational resources are too limited to use a $batch\_size$ value proposed in \cite{sun19}, we set $batch\_size=16$. Another important thing in any task of text classification is tackling the long texts. We have to take into account the fact that both RuBERT and XLM-RoBERTa cannot process texts that are longer than 512 tokens. In our study, we just take the first 512 tokens, also known as \textit{head-only} truncation method from \cite{sun19}.

As a traditional classifier we focus Logistic Regression because it is a traditional ML algorithm which differs drastically from that of both XLM-RoBERTa and RuBERT. Moreover, it is quite a lightweight classifier that has a decent speed of training and inference.
We use 5000 word-based features and 10000 char-based features, all of which are selected by the highest tf-idf value.  The char-based and the word-based features are represented by 2-, 3- and 4-gramms of chars and words correspondingly. 

We train two types of ensembles -- one of them consists of the XLM-RoBERTa and the RuBERT genre classifiers, the second one additionally includes a traditional ML classifier, mostly Logistic Regression, while in some experiments we also ran SVM and RF for comparison with LR. We name them correspondingly \textit{ensemble2} and \textit{ensemble3}.  The ensemble weights are selected independently for each pair (train dataset, test dataset) on the basis of the performance on a validation set. A relatively high value of the LR weights for Ensemble 3 (even though this is not the strongest classifier) comes from the difference in the probability levels, as the median probability for the predicted class is around 0.98 for the transformer classifiers, while it is 0.68 for LR.  Among the transformer classifiers, the XLM-RoBERTa weight is typically higher. This matches our findings in \autoref{tabFMeasure}, where XLM-RoBERTa obtains the highest f1-score among the separate classifiers.

\begin{table*}[!t]
\centering
\begin{tabular}{l|l|llll|llll}
Train & Val/ & \multicolumn{4}{c}{Ensemble2} & \multicolumn{4}{c}{Ensemble3} \\
& Test& XLM-R & RuBERT & LR & Acc & XLM-R & RuBERT & LR & Acc \\
\hline
FTD & FTD & 0.692 & 0.308 & 0 & 0.840 & 0.158 & 0.222 & 0.620 & 0.882 \\
FTD & LJ & 0.821 & 0.179 & 0 & 0.671 & 0.684 & 0.017 & 0.299 & 0.674 \\
LJ & FTD & 0.513 & 0.487 & 0 & 0.701 & 0.474 & 0.139 & 0.388 & 0.708 \\
LJ & LJ & 0.692 & 0.308 & 0 & 0.775 & 0.421 & 0.274 & 0.305 & 0.785 \\
LJ+FTD & FTD & 0.769 & 0.231 & 0 & 0.826 & 0.211 & 0.125 & 0.665 & 0.875 \\
LJ+FTD & LJ & 0.718 & 0.282 & 0 & 0.770 & 0.474 & 0.166 & 0.360 & 0.776 \\
\hline
\end{tabular}
\caption{Ensemble weights \label{tabEnsembleCoefficients}}
\end{table*}

As we can see in \autoref{tabFMeasure}, the f1 metric of each model strongly depends on the genre. Each classifier is worse in predicting informational texts than most other genres. Such a suboptimal result on the informational texts is likely to be caused by their heterogeneity as they include texts of very different classes, such as encyclopedias, dictionaries, CVs, biographies, product specifications, etc. 

\autoref{dependenceOnTrainSize} shows how much the models' accuracy depends on the training data size. The models continue to improve with the increase in training data. We can also see than each ensemble almost always works better than every individual model, regardless of the size of the training data. 

The domain mismatch between the FTD and the LiveJournal corpora is shown in the drop of accuracy when training on FTD and testing on LiveJournal or vice versa \autoref{tabFMeasure}. Each classifier trained on FTD performs much worse on LiveJournal than on FTD. The same situation appears for any classifier trained on LiveJournal. This can be seen clearly for the genre Promotion. The classifiers show relatively high accuracy on the FTD testing promotion texts and a suboptimal one on the promotion texts from LiveJournal. One reason is the topical shift as FTD does not include promotion texts that advertise websites or internet services. Since most of such Promotion texts have numerous keywords on the topic \textit{internet}, a classifier that was not trained on the Promotion texts on this topic is unlikely to identify the genre correctly. We can see, that even the classifiers trained on the LJ training subset do not perform well on the LJ testing data, though the result is not that bad as it is for the FTD-trained models.  It is an example of domain shifts that distort the prediction of genres by models.

In order to balance the mismatch between LiveJournal and FTD, we train the classification architectures on the concatenation of LiveJournal and FTD. As we can see in \autoref{tabFMeasureConcat}, these classifiers perform much better on LiveJournal than those trained only on FTD. Similarly, the classifiers are more accurate on FTD than those trained solely on LiveJournal. It shows that including texts from multiple domains into the training set helps to cope with topical shifts in the test.

\begin{table*}[!t]
\setlength{\tabcolsep}{3pt}
\begin{tabular}{llllllll}
Train & Test & Genre       & XLM-R                    & RuBERT                         & LR & Ensemble2                      & Ensemble3                       \\
\hline
FTD & FTD & Argument & \textbf{0.729}$^{\pm 0.021}$ & \textbf{0.695}$^{\pm 0.027}$ & \textbf{0.721} & \textbf{0.744}$^{\pm 0.016}$ & \textbf{0.775}$^{\pm 0.016}$\\
FTD & FTD & Fiction & \textbf{0.690}$^{\pm 0.051}$ & \textbf{0.728}$^{\pm 0.043}$ & \textbf{0.789} & \textbf{0.757}$^{\pm 0.058}$ & \textbf{0.795}$^{\pm 0.030}$\\
FTD & FTD & Instruction & \textbf{0.760}$^{\pm 0.104}$ & \textbf{0.790}$^{\pm 0.049}$ & \textbf{0.741} & \textbf{0.811}$^{\pm 0.060}$ & \textbf{0.810}$^{\pm 0.039}$\\
FTD & FTD & News & \textbf{0.944}$^{\pm 0.011}$ & \textbf{0.920}$^{\pm 0.014}$ & 0.904 & \textbf{0.937}$^{\pm 0.016}$ & \textbf{0.936}$^{\pm 0.009}$\\
FTD & FTD & Legal & 0.757$^{\pm 0.039}$ & 0.793$^{\pm 0.043}$ & \textbf{0.923} & 0.789$^{\pm 0.042}$ & 0.826$^{\pm 0.041}$\\
FTD & FTD & Personal & \textbf{0.725}$^{\pm 0.028}$ & \textbf{0.698}$^{\pm 0.030}$ & \textbf{0.708} & \textbf{0.732}$^{\pm 0.034}$ & \textbf{0.750}$^{\pm 0.024}$\\
FTD & FTD & Promotion & \textbf{0.937}$^{\pm 0.012}$ & 0.918$^{\pm 0.008}$ & \textbf{0.924} & \textbf{0.939}$^{\pm 0.010}$ & \textbf{0.937}$^{\pm 0.006}$\\
FTD & FTD & Academic & \textbf{0.883}$^{\pm 0.023}$ & \textbf{0.876}$^{\pm 0.022}$ & 0.820 & \textbf{0.892}$^{\pm 0.018}$ & \textbf{0.886}$^{\pm 0.014}$\\
FTD & FTD & Information & \textbf{0.657}$^{\pm 0.047}$ & \textbf{0.587}$^{\pm 0.075}$ & 0.293 & \textbf{0.670}$^{\pm 0.054}$ & \textbf{0.654}$^{\pm 0.047}$\\
FTD & FTD & Review & 0.711$^{\pm 0.030}$ & 0.698$^{\pm 0.037}$ & \textbf{0.763} & 0.743$^{\pm 0.019}$ & \textbf{0.782}$^{\pm 0.022}$\\
\hline
FTD & LJ & Argument & \textbf{0.475}$^{\pm 0.019}$ & \textbf{0.425}$^{\pm 0.037}$ & \textbf{0.429} & \textbf{0.477}$^{\pm 0.021}$ & \textbf{0.480}$^{\pm 0.020}$\\
FTD & LJ & Fiction & \textbf{0.675}$^{\pm 0.033}$ & 0.605$^{\pm 0.050}$ & 0.346 & \textbf{0.697}$^{\pm 0.033}$ & \textbf{0.699}$^{\pm 0.035}$\\
FTD & LJ & Instruction & \textbf{0.539}$^{\pm 0.084}$ & \textbf{0.534}$^{\pm 0.036}$ & 0.381 & \textbf{0.548}$^{\pm 0.064}$ & \textbf{0.544}$^{\pm 0.061}$\\
FTD & LJ & News & \textbf{0.877}$^{\pm 0.010}$ & 0.853$^{\pm 0.008}$ & 0.745 & \textbf{0.876}$^{\pm 0.010}$ & \textbf{0.876}$^{\pm 0.010}$\\
FTD & LJ & Legal & \textbf{0.495}$^{\pm 0.079}$ & 0.415$^{\pm 0.077}$ & 0.563 & \textbf{0.495}$^{\pm 0.072}$ & 0.489$^{\pm 0.059}$\\
FTD & LJ & Personal & \textbf{0.716}$^{\pm 0.027}$ & 0.685$^{\pm 0.021}$ & 0.616 & \textbf{0.727}$^{\pm 0.015}$ & \textbf{0.730}$^{\pm 0.010}$\\
FTD & LJ & Promotion & \textbf{0.255}$^{\pm 0.046}$ & \textbf{0.252}$^{\pm 0.024}$ & 0.209 & \textbf{0.265}$^{\pm 0.036}$ & \textbf{0.265}$^{\pm 0.027}$\\
FTD & LJ & Academic & \textbf{0.111}$^{\pm 0.063}$ & \textbf{0.122}$^{\pm 0.048}$ & \textbf{0.105} & \textbf{0.151}$^{\pm 0.042}$ & \textbf{0.127}$^{\pm 0.037}$\\
FTD & LJ & Information & \textbf{0.541}$^{\pm 0.027}$ & 0.364$^{\pm 0.129}$ & 0.093 & \textbf{0.534}$^{\pm 0.043}$ & \textbf{0.520}$^{\pm 0.050}$\\
FTD & LJ & Review & \textbf{0.486}$^{\pm 0.044}$ & 0.443$^{\pm 0.031}$ & 0.341 & \textbf{0.495}$^{\pm 0.026}$ & \textbf{0.501}$^{\pm 0.023}$\\

\hline

LJ & FTD & Argument & \textbf{0.565}$^{\pm 0.027}$ & \textbf{0.589}$^{\pm 0.045}$ & \textbf{0.550} & \textbf{0.594}$^{\pm 0.029}$ & \textbf{0.594}$^{\pm 0.029}$\\
LJ & FTD & Fiction & \textbf{0.710}$^{\pm 0.043}$ & \textbf{0.774}$^{\pm 0.038}$ & \textbf{0.739} & \textbf{0.770}$^{\pm 0.034}$ & \textbf{0.770}$^{\pm 0.034}$\\
LJ & FTD & Instruction & \textbf{0.493}$^{\pm 0.067}$ & \textbf{0.472}$^{\pm 0.105}$ & 0.371 & \textbf{0.505}$^{\pm 0.068}$ & \textbf{0.505}$^{\pm 0.068}$\\
LJ & FTD & News & \textbf{0.866}$^{\pm 0.017}$ & \textbf{0.857}$^{\pm 0.034}$ & 0.770 & \textbf{0.857}$^{\pm 0.011}$ & \textbf{0.857}$^{\pm 0.011}$\\
LJ & FTD & Legal & \textbf{0.402}$^{\pm 0.305}$ & \textbf{0.518}$^{\pm 0.242}$ & 0.000 & \textbf{0.529}$^{\pm 0.298}$ & \textbf{0.529}$^{\pm 0.298}$\\
LJ & FTD & Personal & \textbf{0.632}$^{\pm 0.037}$ & \textbf{0.610}$^{\pm 0.039}$ & \textbf{0.635} & \textbf{0.646}$^{\pm 0.019}$ & \textbf{0.646}$^{\pm 0.019}$\\
LJ & FTD & Promotion & \textbf{0.782}$^{\pm 0.090}$ & \textbf{0.728}$^{\pm 0.104}$ & 0.455 & \textbf{0.790}$^{\pm 0.076}$ & \textbf{0.790}$^{\pm 0.076}$\\
LJ & FTD & Academic & \textbf{0.249}$^{\pm 0.218}$ & \textbf{0.300}$^{\pm 0.182}$ & 0.000 & \textbf{0.270}$^{\pm 0.138}$ & \textbf{0.270}$^{\pm 0.138}$\\
LJ & FTD & Information & \textbf{0.461}$^{\pm 0.037}$ & \textbf{0.444}$^{\pm 0.034}$ & 0.321 & \textbf{0.462}$^{\pm 0.033}$ & \textbf{0.462}$^{\pm 0.033}$\\
LJ & FTD & Review & \textbf{0.571}$^{\pm 0.054}$ & \textbf{0.492}$^{\pm 0.075}$ & \textbf{0.418} & \textbf{0.569}$^{\pm 0.042}$ & \textbf{0.569}$^{\pm 0.042}$\\
\hline
LJ & LJ & Argument & \textbf{0.591}$^{\pm 0.021}$ & \textbf{0.580}$^{\pm 0.026}$ & 0.500 & \textbf{0.605}$^{\pm 0.021}$ & \textbf{0.605}$^{\pm 0.021}$\\
LJ & LJ & Fiction & \textbf{0.769}$^{\pm 0.018}$ & \textbf{0.740}$^{\pm 0.020}$ & 0.637 & \textbf{0.776}$^{\pm 0.012}$ & \textbf{0.776}$^{\pm 0.012}$\\
LJ & LJ & Instruction & \textbf{0.804}$^{\pm 0.014}$ & \textbf{0.794}$^{\pm 0.013}$ & 0.768 & \textbf{0.815}$^{\pm 0.008}$ & \textbf{0.815}$^{\pm 0.008}$\\
LJ & LJ & News & \textbf{0.912}$^{\pm 0.006}$ & 0.899$^{\pm 0.004}$ & 0.864 & \textbf{0.912}$^{\pm 0.005}$ & \textbf{0.912}$^{\pm 0.005}$\\
LJ & LJ & Legal & \textbf{0.172}$^{\pm 0.165}$ & \textbf{0.236}$^{\pm 0.221}$ & 0.154 & \textbf{0.153}$^{\pm 0.140}$ & \textbf{0.153}$^{\pm 0.140}$\\
LJ & LJ & Personal & \textbf{0.813}$^{\pm 0.012}$ & \textbf{0.795}$^{\pm 0.011}$ & 0.742 & \textbf{0.820}$^{\pm 0.008}$ & \textbf{0.820}$^{\pm 0.008}$\\
LJ & LJ & Promotion & \textbf{0.480}$^{\pm 0.059}$ & \textbf{0.455}$^{\pm 0.048}$ & 0.289 & \textbf{0.491}$^{\pm 0.051}$ & \textbf{0.491}$^{\pm 0.051}$\\
LJ & LJ & Academic & \textbf{0.378}$^{\pm 0.187}$ & \textbf{0.376}$^{\pm 0.142}$ & 0.000 & \textbf{0.361}$^{\pm 0.138}$ & \textbf{0.361}$^{\pm 0.138}$\\
LJ & LJ & Information & \textbf{0.667}$^{\pm 0.013}$ & \textbf{0.664}$^{\pm 0.019}$ & 0.539 & \textbf{0.681}$^{\pm 0.010}$ & \textbf{0.681}$^{\pm 0.010}$\\
LJ & LJ & Review & \textbf{0.641}$^{\pm 0.017}$ & \textbf{0.622}$^{\pm 0.032}$ & 0.520 & \textbf{0.665}$^{\pm 0.019}$ & \textbf{0.665}$^{\pm 0.019}$\\
\hline
\end{tabular}
\caption{F1-measures for classifiers trained and tested on different datasets}
\label{tabFMeasure}
\end{table*}

\begin{table*}[!t]
\centering
\begin{tabular}{l|rr|rr|rr}
Dataset & \multicolumn{2}{c}{XLM-R} & \multicolumn{2}{c}{RuBERT} & \multicolumn{2}{c}{LR} \\
             & train &  test & train & test & train & test \\
\hline
FTD & 101.3 & 34.7 & 96.7 & 34.7 & 18.8 & 0.1 \\
LJ & 115 & 51.1 & 101.5 & 51 & 8.7 & 0.1 \\
\hline
\end{tabular}
\caption{Training and inference time, milliseconds averaged per text \label{tabTime}}
\end{table*}

To train each neural model in the experiments, we use single NVIDIA TITAN RTX.
\autoref{tabTime} shows that the training and inference time for XLM-RoBERTa and RuBERT are at the same level, whilst the Logistic Regression is much faster to train and test per text. 

\begin{table*}[!t]
\setlength{\tabcolsep}{3pt}
\begin{tabular}{llllllll}
Train & Test & Genre       & XLM-R                    & RuBERT                         & LR & Ensemble2                      & Ensemble3                       \\
\hline
LJ+FTD & FTD & Argument & \textbf{0.708}$^{\pm 0.030}$ & 0.693$^{\pm 0.022}$ & 0.550 & \textbf{0.732}$^{\pm 0.012}$ & \textbf{0.716}$^{\pm 0.022}$\\
LJ+FTD & FTD & Fiction & \textbf{0.780}$^{\pm 0.047}$ & \textbf{0.784}$^{\pm 0.053}$ & 0.739 & \textbf{0.809}$^{\pm 0.034}$ & \textbf{0.807}$^{\pm 0.017}$\\
LJ+FTD & FTD & Instruction & \textbf{0.761}$^{\pm 0.062}$ & 0.665$^{\pm 0.037}$ & 0.371 & \textbf{0.742}$^{\pm 0.034}$ & \textbf{0.714}$^{\pm 0.039}$\\
LJ+FTD & FTD & News & \textbf{0.945}$^{\pm 0.008}$ & \textbf{0.929}$^{\pm 0.012}$ & 0.770 & \textbf{0.945}$^{\pm 0.011}$ & 0.913$^{\pm 0.020}$\\
LJ+FTD & FTD & Legal & \textbf{0.795}$^{\pm 0.081}$ & \textbf{0.773}$^{\pm 0.037}$ & 0.000 & \textbf{0.788}$^{\pm 0.056}$ & \textbf{0.750}$^{\pm 0.133}$\\
LJ+FTD & FTD & Personal & \textbf{0.714}$^{\pm 0.024}$ & \textbf{0.661}$^{\pm 0.041}$ & 0.635 & \textbf{0.711}$^{\pm 0.027}$ & \textbf{0.713}$^{\pm 0.018}$\\
LJ+FTD & FTD & Promotion & \textbf{0.946}$^{\pm 0.011}$ & 0.908$^{\pm 0.018}$ & 0.455 & \textbf{0.946}$^{\pm 0.015}$ & \textbf{0.937}$^{\pm 0.010}$\\
LJ+FTD & FTD & Academic & \textbf{0.880}$^{\pm 0.015}$ & \textbf{0.852}$^{\pm 0.056}$ & 0.000 & \textbf{0.892}$^{\pm 0.017}$ & \textbf{0.852}$^{\pm 0.041}$\\
LJ+FTD & FTD & Information & \textbf{0.650}$^{\pm 0.053}$ & \textbf{0.627}$^{\pm 0.038}$ & 0.321 & \textbf{0.670}$^{\pm 0.043}$ & \textbf{0.624}$^{\pm 0.036}$\\
LJ+FTD & FTD & Review & \textbf{0.685}$^{\pm 0.041}$ & 0.590$^{\pm 0.048}$ & 0.418 & \textbf{0.687}$^{\pm 0.048}$ & \textbf{0.653}$^{\pm 0.028}$\\
\hline
LJ+FTD & LJ & Argument & \textbf{0.590}$^{\pm 0.020}$ & \textbf{0.563}$^{\pm 0.035}$ & 0.500 & \textbf{0.603}$^{\pm 0.023}$ & \textbf{0.609}$^{\pm 0.012}$\\
LJ+FTD & LJ & Fiction & \textbf{0.734}$^{\pm 0.028}$ & \textbf{0.716}$^{\pm 0.026}$ & 0.637 & \textbf{0.756}$^{\pm 0.028}$ & \textbf{0.762}$^{\pm 0.023}$\\
LJ+FTD & LJ & Instruction & \textbf{0.795}$^{\pm 0.021}$ & 0.787$^{\pm 0.011}$ & 0.768 & \textbf{0.810}$^{\pm 0.014}$ & \textbf{0.818}$^{\pm 0.011}$\\
LJ+FTD & LJ & News & \textbf{0.907}$^{\pm 0.006}$ & \textbf{0.904}$^{\pm 0.007}$ & 0.864 & \textbf{0.912}$^{\pm 0.005}$ & \textbf{0.914}$^{\pm 0.004}$\\
LJ+FTD & LJ & Legal & \textbf{0.331}$^{\pm 0.216}$ & \textbf{0.394}$^{\pm 0.156}$ & 0.154 & \textbf{0.355}$^{\pm 0.143}$ & \textbf{0.332}$^{\pm 0.126}$\\
LJ+FTD & LJ & Personal & \textbf{0.807}$^{\pm 0.011}$ & 0.783$^{\pm 0.018}$ & 0.742 & \textbf{0.816}$^{\pm 0.010}$ & \textbf{0.817}$^{\pm 0.004}$\\
LJ+FTD & LJ & Promotion & \textbf{0.465}$^{\pm 0.049}$ & \textbf{0.479}$^{\pm 0.047}$ & 0.289 & \textbf{0.495}$^{\pm 0.038}$ & \textbf{0.488}$^{\pm 0.051}$\\
LJ+FTD & LJ & Academic & \textbf{0.324}$^{\pm 0.104}$ & \textbf{0.332}$^{\pm 0.107}$ & 0.000 & \textbf{0.409}$^{\pm 0.084}$ & \textbf{0.350}$^{\pm 0.158}$\\
LJ+FTD & LJ & Information & 0.642$^{\pm 0.014}$ & \textbf{0.641}$^{\pm 0.030}$ & 0.539 & \textbf{0.658}$^{\pm 0.018}$ & \textbf{0.674}$^{\pm 0.012}$\\
LJ+FTD & LJ & Review & \textbf{0.640}$^{\pm 0.017}$ & 0.604$^{\pm 0.026}$ & 0.520 & \textbf{0.653}$^{\pm 0.015}$ & \textbf{0.658}$^{\pm 0.016}$\\
\hline
LJ+FTD & FTD & Total accuracy & \textbf{0.822}$^{\pm 0.005}$ & 0.789$^{\pm 0.009}$ & 0.532 & \textbf{0.828}$^{\pm 0.007}$ & \textbf{0.828}$^{\pm 0.007}$ \\
LJ+FTD & LJ & Total accuracy & \textbf{0.757}$^{\pm 0.011}$ & \textbf{0.741}$^{\pm 0.010}$ & 0.694 & \textbf{0.769}$^{\pm 0.010}$ & \textbf{0.774}$^{\pm 0.006}$ \\

\end{tabular}
\caption{F1-measures for classifiers trained on the concatenation of FTD and LJ}
\label{tabFMeasureConcat}
\end{table*}


\autoref{tabFMeasure} also show how much the transformer classifiers depend on the random seed during model training, so we report the confidence intervals over ten random seeds and we highlight in bold the best F-scores if there is no other confidence interval that lies entirely to the right of it without intersection. Often the intersection of the confidence intervals for the best F-scores are quite frequent.

In most cases, the ensemble attains the highest f1-measure. Among the individual classifiers, XLM-RoBERTa seems to be the most reliable one. \autoref{tabFMeasure} shows that the number of genres for which XLM-RoBERTa is the best is lower than that for the ensemble but higher than the values for the RuBERT and the Logistic Regression genre classifiers. 

We also tried to replace Logistic Regression with other weak classifiers, including SVM (with standard hyperparameters from sklearn) with linear and RBF kernels and Random Forest (with 100 and 1000 trees). The ensemble with linear SVM attains total accuracy $0.828 \pm 0.004$ on FTD and $0.771 \pm 0.009$ on LJ.
The ensemble with Random Forest attains total accuracy $828 \pm 0.004$ on FTD and $0.770 \pm 0.011$ on LJ. The Random Forest classifier of 1000 trees does not increase accuracy compared to Random Forest of 100 trees but it requires 10 times more time for training and testing.  Ensembles with them show higher accuracy than that of individual XLM-RoBERTa and RuBERT but lower than the ensemble with Logistic Regression. Moreover, the training and inference time of SVM and Random Forest turn out to be much higher than that for Logistic Regression. These results support choosing Logistic Regression as a classifier for further experiments.

\begin{table*}[]
\centering
\begin{tabular}{l|l|p{2.7cm}|p{2.7cm}|p{2.7cm}|p{2.7cm}}
Train & Test & XLM-R                    & RuBERT                         & LogReg                       & Ensemble3            \\
\hline
FTD & FTD & (A4, A1, 0.217) \newline (A16, A1, 0.212) \newline (A11, A1, 0.122) \newline (A16, A9, 0.121) & (A4, A1, 0.130) \newline (A16, A14, 0.121) \newline (A16, A17, 0.121) \newline (A11, A17, 0.102) & (A16, A1, 0.303) \newline (A4, A1, 0.217) \newline (A16, A8, 0.182) \newline (A1, A17, 0.163) \newline (A16, A14, 0.152) & (A16, A1, 0.242) \newline (A4, A1, 0.217) \newline (A4, A11, 0.130) \newline(A17, A11, 0.118) \\   
\hline
FTD & LJ & (A14, A1, 0.5) \newline (A4, A11, 0.308) \newline (A1, A11, 0.305) \newline (A7, A1, 0.288) \newline (A4, A1, 0.269) & (A14, A1, 0.5) \newline (A1, A11, 0.274) \newline (A4, A1, 0.225) \newline (A4, A12, 0.225) \newline (A16, A12, 0.211) & (A4, A11, 0.615) \newline (A14, A8, 0.5) \newline (A16, A8, 0.421) \newline (A7, A12, 0.338) \newline (A17, A11, 0.297) & (A14, A1, 0.5) \newline (A4, A11, 0.346)  \newline (A1, A11, 0.263) \newline (A7, A1, 0.25) \newline (A16, A12, 0.237) \\   
\hline
LJ & FTD & (A16, A1, 0.242) \newline (A17, A11, 0.206) \newline (A11, A1, 0.163) \newline (A4, A11, 0.130) \newline (A7, A17, 0.118) & (A17, A11, 0.206) \newline (A11, A1, 0.184) \newline (A9, A7, 0.154) \newline (A16, A1, 0,152) \newline (A17, A1, 0.147) & (A16, A1, 0.242) \newline (A4, A11, 0.217) \newline (A17, A11, 0.206) \newline (A11, A1, 0.184) \newline (A17, A1, 0.147) & (A16, A1, 0.212) \newline (A17, A11, 0.206) \newline (A11, A1, 0.163) \newline (A4, A11, 0.130) \newline (A7, A17, 0.118) \\   
\hline
LJ & LJ & (A14, A1, 0.5) \newline (A12, A7, 0.286) \newline (A12, A8, 0.286) \newline (A1, A11, 0.190) \newline (A12, A17, 0.143) & (A14, A1, 0.5) \newline (A12, A17, 0.286) \newline (A1, A11, 0.211) \newline (A17, A11, 0.162) \newline (A4, A1, 0.154) & (A9, A7, 1.000) \newline (A12, A7, 0.571) \newline (A14, A1, 0.500) \newline (A17, A11, 0.324) \newline (A12, A16, 0.286) & (A14, A1, 0.500) \newline (A12, A7, 0.286) \newline (A12, A8, 0.286) \newline (A17, A11, 0.270) \newline (A4, A1, 0.192) \\
\hline

\end{tabular}
\caption{The most common items in the confusion matrices}
\label{tabConfusionPairs}
\end{table*}

The most frequent kinds of mistakes are shown in \autoref{tabConfusionPairs}. We show the percentage of the texts in the test data that are misclassified in a certain way. In the table, we use the shortcuts for the genres as per \autoref{tabTraining}. Surprisingly, confusion pairs do not vary too much with changing of genre classifier. Besides, the nature of the mistakes is quite predictable and it is related to the similarities between the genres, often leading to genre hybrids. Functions of some Argumentation and Informational texts might be similar. The same can be told about genre pairs (Review, Personal), (Research, Argumentation), (Info, Research), etc. The most important difference between the confusion matrices of different genre classifiers is the relative number of mistakes. For example, the Ensemble3 and the XLM-RoBERTa show lower confusion numbers in \autoref{tabConfusionPairs}.

\subsection{Optimal hyperparameters}
\label{subsecOptimalEpochs}

\begin{figure}[!t]
\centering
  \includegraphics[width=.9\linewidth]{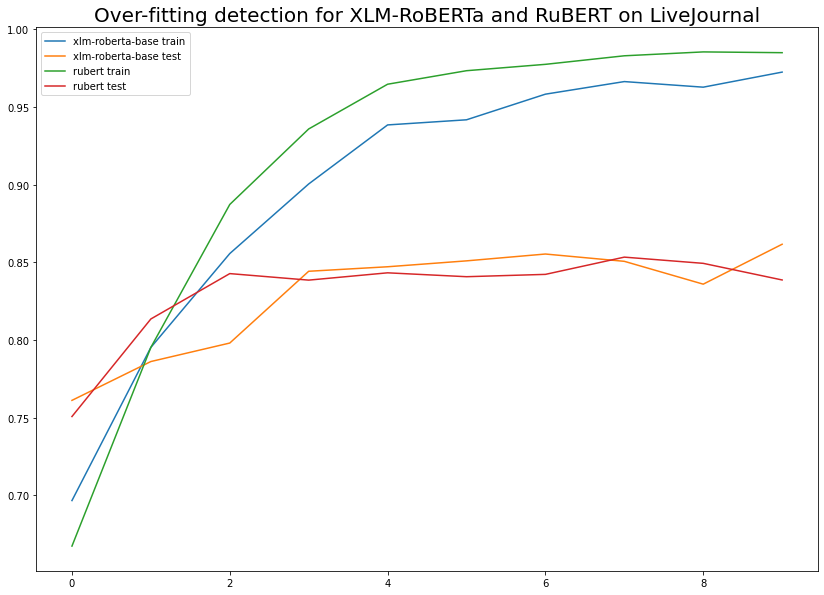}
  \caption{Learning curve of xlm-roberta-base and rubert-base-cased on LiveJournal}
  \label{learningCurveLJ}
\end{figure}

XLM-RoBERTa and RuBERT start to over-fit after 3 epochs, see \autoref{learningCurveLJ} for the LJ corpus, the behaviour for FTD is similar. Since the third epoch, RuBERT continues improving on the training subset, while its accuracy on the test does not change. It looks like RuBERT is slightly more vulnerable to over-fitting beyond 3-4 epochs on the genre classification task. The authors of \cite{sun19} claim that for their data and BERT-like models, the optimal number of epochs equals 4 on their text classification tasks. Since \cite{sun19} do not use RuBERT or XLM-RoBERTa, the novelty of our study is also in the fact that we show optimality of this number of epochs for other BERT-like models.

\subsection{Experiments on the big datasets}
\label{subsecBigDatasets}

We apply the classifiers to large social media samples from VKontakte and LJ datasets from the General Internet Corpus of Russian \cite{piperski13} to get an estimate of the genre distributions in social media. \autoref{tabBigLJ} and \autoref{tabBigVK} list the corresponding distributions of the predicted classes.  The most frequent class in both cases is Personal reporting with the second most common genre is Argumentation, while there are far fewer Legal and Academic texts. This result correlates well with the expected distribution of texts in social media and gives some validation for the accuracy of our predictions in the absence of a very large test set. We use chi-squared test to find the most significant differences in the shares of genres in these two samples. The LiveJournal sample has a bigger share of Legal and Argumentative texts, as it is often used as a blogging platform rather than the more traditional social network of VKontakte. In turn, VKontakte has a significantly higher share of Personal reporting as well as Promotional texts in comparison to LJ. 
 
\begin{table*}[!t]
\centering
\begin{tabular}{lrrrrrrrrr}
Genre label & XLM-R & RuBERT  & LogReg & Ensemble3 & Percentage \\
\hline
Argument 	& 11192 	& 21568 & 11773 & 15454 & 20.8 \\
Fiction 	& 5182	& 6801  & 5471 & 5836 & 7.86  \\
Instruction & 5583	& 3635  & 3465 & 3531 & 4.76 \\
News    	& 6925 	& 7548  & 11079 & 8674 & 11.68 \\
Legal   	& 33 	& 133   & 23 & 83 & 0.11 \\
Personal 	& 34127 	& 24427 & 37315 & 32099 & 43.22 \\
Promotion 	& 2722  	& 2548  & 1489 & 2059 & 2.77 \\
Academic 	& 67 	& 126   & 30 & 74 & 0.1 \\
Information & 1118   & 1337  & 1214 & 1156 & 1.56 \\
Review   	& 5666	& 6009  & 2408 & 4021 & 5.41 \\
Non-text       & 384    & 134   &    0.04   &  10 & 0.01 \\
\hline
   \textbf{Total} & 74267 & 74267 & 74267 & 74267 & 100 \\
\end{tabular}
\caption{Prediction distribution on the LJ sample, divided by $10^3$ 
\label{tabBigLJ}}
\end{table*}

\begin{table*}[!t]
\centering
\begin{tabular}{lrrrrrrrrr}
Genre label & XLM-R & RuBERT  & LogReg & Ensemble3 &  Percentage \\
\hline
Argument 	& 19941	& 46514 & 20344 & 30325 & 13.63 \\
Fiction 	& 5182	& 33320  & 22780 & 26917 & 12.01 \\
Instruction & 5583	& 12706  & 7960 & 9579 & 4.31 \\
News    	& 6925 	& 8841  & 28150 & 12487 & 5.61 \\
Legal   	& 33 	& 168   & 45 & 86 & 0.04 \\
Personal 	& 123338 & 89444 & 124520 & 116983 &  52.59 \\
Promotion 	& 25816  & 17604  & 13073 & 17260 & 7.76 \\
Academic 	& 284 	& 249   & 131 & 186 &  0.08 \\
Information & 2341   & 2106  & 931 & 1397 & 0.62 \\
Review   	& 9859	& 11351  & 4526 & 7214 & 3.24 \\
Non-text       & 481    & 156   &    0.015   &   24 & 0.01 \\

\hline
   \textbf{Total} & 222459 & 222459 & 222459 & 222459 & 100 \\
\end{tabular}
\caption{Prediction distribution on the VKontakte sample, divided by $10^3$
\label{tabBigVK}}
\end{table*}

\section{Experiment on the confidence of predictions}
\label{secConfidence}

The evaluation of confidence depends significantly on text genre and hence requires a lot of texts for each genre for getting a stable estimation. For this reason, we apply only those classifiers which are trained in the concatenation of the LiveJournal and FTD training corpora, while also testing their concatenation.

\begin{table*}[!t]
\centering
\begin{tabular}{llllllllllll}
Genre &
\multicolumn{2}{c}{XLM-RoBERTa} & \multicolumn{2}{c}{RuBERT} & \multicolumn{2}{c}{LogReg} & 
\multicolumn{2}{c}{Ensemble 2} &
\multicolumn{2}{c}{Ensemble 3}
\\
     & stat & delta & stat & delta & stat & delta & stat & delta & stat & delta \\
\hline
Argument & \textbf{0.697} & \textbf{0.123} & 0.624 & 0.08 & 0.596 & 0.052 & 0.67 & 0.101 & 0.638 & 0.075 \\
Fiction & 0.705 & 0.130 & \textbf{0.749} & 0.148 & 0.635 & 0.071 & 0.732 & \textbf{0.158} & 0.716 & 0.126 \\
Instruction & 0.838 & 0.224 & 0.800 & 0.179 & \textbf{0.878} & \textbf{0.254} & 0.834 & 0.229 & 0.845 & 0.236 \\
News & 0.914 & 0.282 & 0.919 & 0.225 & 0.909 & 0.325 & 0.898 & 0.254 & \textbf{0.929} & \textbf{0.335} \\
Legal & 0.740 & 0.145 & 0.728 & 0.119 & 0.513 & 0.016 & 0.743 & 0.149 & \textbf{0.763} & \textbf{0.153} \\
Personal & 0.830 & 0.193 & \textbf{0.897} & 0.233 & 0.848 & 0.213 & 0.858 & \textbf{0.234} & 0.863 & 0.228 \\
Promotion & 0.878 & 0.309 & 0.878 & 0.282 & 0.870 & 0.254 & 0.899 & 0.336 & \textbf{0.915} & \textbf{0.349} \\
Academic & 0.854 & \textbf{0.278} & 0.786 & 0.125 & \textbf{0.885} & 0.249 & 0.778 & 0.17 & 0.828 & 0.243 \\
Information & 0.683 & 0.109 & 0.685 & 0.124 & 0.586 & 0.044 & \textbf{0.721} & \textbf{0.141} & 0.648 & 0.079 \\
Review & \textbf{0.690} & \textbf{0.118} & 0.557 & 0.038 & 0.576 & 0.047 & 0.641 & 0.094 & 0.640 & 0.080 \\
\hline
Total & \textbf{0.816} & 0.211 & 0.808 & 0.183 & 0.792 & 0.199 & 0.815 & 0.214 & 0.815 & \textbf{0.217}
\end{tabular}
\caption{Mann-Whitney test for the confidence of the genre classifiers
\label{tabConfidence}}
\end{table*}

\autoref{tabConfidence} shows the confidence delta in the cases when the classifiers make a correct prediction, and when they make a mistake. The delta value is positive, which indicates that the higher confidence does correspond to a correct prediction. 

We also conducted an unpaired two-sample Mann-Whitney test for each classifier and genre for the confidence levels for the correct and incorrect predictions. For 3 genres of 10, Ensemble3 achieves the highest value of Mann-Whitney statistics and confidence delta, while it is nearly always at least the second best. Moreover, it has the highest delta on the entire testing subset. The genre classifier based on Logistic Regression performs the worst.

\section{Related Work}
\label{secRelatedWork}

Genre classification is not a new task. Up to date, a lot of attempts have appeared to build a precise classifier of genres based on various architectures from linear discrimination \cite{karlgren94} to SVM \cite{sharoff10lrec} and recurrent neural networks \cite{kunilovskaya19}. Our study differs in the way that instead of using classical ML techniques, we apply advanced transformer-based neural architectures -- XLM-RoBERTa and RuBERT.   \cite{ronnqvist21multilingual} produced a closely-related experiment which involved fine-tuning an XLM-RoBERTa model to predict genres for a range of languages. Among other things they showed the multilingual potential of XLM-RoBERTa, a genre classifier can be trained on one language and tested on another one. However, they have not tested robustness of predictions.

\cite{sharoff10lrec} contains a study on the performance of SVM for the problem of genre classification. \cite{sharoff10lrec} and \cite{kunilovskaya19} use the Russian FTD corpus for training genre classifiers. In our study, we use the same FTD corpus. The validation accuracy in both \cite{sharoff10lrec} and \cite{kunilovskaya19} is significantly lower than in our paper.
Our work is different in that we use the LiveJournal corpus to validate our models by taking into account the topical shifts.

\cite{dropout_strikes_back} describes the method to measure prediction confidence that we use in our study. The key idea of this metric presumes that a classifier can only be confident on some text examples if it predicts the correct labels to the texts with similar embeddings. We do not bring any change to the definition and the algorithm of calculation of confidence of prediction. Our major contribution is the application of this metric to the task of genre segmentation. Moreover, our study reaffirms the result of \cite{dropout_strikes_back} that ensembles are more confident in their predictions, since the confidence delta for the ensembles is higher than that for each individual classifier.

In the original paper on BERT \cite{bert}, there are some tips for training and fine-tuning the BERT architecture for text classification given by the authors. The paper recommends using 2-4 epochs for training or fine-tuning and claims that using more epochs leads to over-fitting. It coincides completely with our experiment on the optimal number of epochs. But, since our training and testing data are in Russian, we use RuBERT and XLM-RoBERTa instead of the base English BERT. Thus, we manage to approve the pieces of advice from \cite{bert} and show that they are valid for Russian and multilingual BERT-based models.

\cite{sun19} is also an important work on how to apply the BERT architecture to the task of text classification.  It researches how multiple hyperparameters for training BERT affect its performance on various Natural Language Understanding tasks. The authors propose 3 different text truncation strategies - \textit{head-only}, \textit{tail-only}, \textit{head-only}, and \textit{head+tail}. Their experiment shows that the \textit{head+tail} method attains the best result. But we decide to use \textit{head-only} instead of it since it is not that obvious how the \textit{head+tail} method affects the topical shifts in data. Another hyperparameter we change is $batch\_size$ due to our limited computational resources. All the other recommendations from the paper are followed.

\section{Conclusions and future research}
\label{secConclusion}

We show that:
\begin{enumerate}[noitemsep]
\item The transformer-based classifiers (XLM-RoBERTa or RuBert) are generally accurate in non-topical classification tasks provided that enough training data is available for each label.
\item For most genres, the ensembles of several classifiers obtain a higher f1-score than any of the separated classifiers.
\item Adding even a weaker classifier, in our case, Logistic Regression, to the ensemble does benefit the classification accuracy. Moreover, the training and inference time for logistic regression is insignificant compared to the corresponding costs for XLM-RoBERTa and RuBERT. It means that adding a weak classifier is worth it.
\item The ensembles have more reliable predictions in terms of their confidence, as they provide the biggest confidence gap between the correctly and incorrectly classified texts.
\item Mann-Whitney statistics shows that the ensemble with Logistic Regression is more reliable for most genres than the ensemble of the two models and each individual classifier.
\item Applying of the classifiers to large social media sample reveals their distribution of genres. Using chi-squared test, we reveal how the genre distributions vary in texts from different social media sources, such as the greater rate of Argumentative texts in LJ in comparison to the greater rate of Personal reporting in VK.
\end{enumerate}

Thus, using ensembles is helpful to boost the accuracy and robustness of predictions while their only disadvantage concerns an increase in the computing costs.

Though ensembles are more reliable for text genre classification, it is still unclear how efficient they are on related tasks.
In future, we consider training models for seq2seq genre segmentation when we need to detect genre boundaries within a single text, for example, when a personal story includes a quote from fiction or when an expression of opinions cites informative news reports. We hypothesise that the ensembles will confirm their advantage over individual models.   The confidence measure can be used to make predictions for open-set genre classification tasks \cite{pritsos18}, when the classifier should refrain from making any prediction for a text, if its confidence on this text is lower than a threshold. 

\section{Bibliographical References}\label{reference}

\bibliographystyle{lrec2022-bib}
\bibliography{bibexport}

\end{document}